\documentclass[journal]{IEEEtran}
\usepackage{cite}
\usepackage{amssymb,amsfonts}
\usepackage{algorithmic}
\usepackage{graphicx}
\usepackage{textcomp}

\usepackage[utf8]{inputenc} 
\usepackage[T1]{fontenc}    
\usepackage{hyperref}       
\usepackage{url}            
\usepackage{booktabs}       
\usepackage{amsfonts}       
\usepackage{nicefrac}       
\usepackage{microtype}      
\usepackage{xcolor}         
\usepackage[linesnumbered,ruled,vlined]{algorithm2e}

\usepackage{amsmath}
\usepackage{amssymb}
\usepackage{latexsym}
\usepackage{graphicx}
\usepackage{multirow}
\usepackage{adjustbox}
\usepackage{makecell}

\usepackage{mathrsfs}

\newcommand{\Bc}{{\mathcal{B}}}
\newcommand{\Cc}{{\mathcal{C}}}
\newcommand{\Hc}{{\mathcal{H}}}
\newcommand{\Tc}{{\mathcal{T}}}

\def\BibTeX{{\rm B\kern-.05em{\sc i\kern-.025em b}\kern-.08em
    T\kern-.1667em\lower.7ex\hbox{E}\kern-.125emX}}

\begin{document}

\title{Multi-Task Distributed Learning using
Vision Transformer with Random Patch Permutation}

\author{Sangjoon Park,  and Jong Chul Ye, \IEEEmembership{Fellow, IEEE}
\thanks{Submitted: April 5, 2022}
\thanks{Sangjoon Park is with the Department of Bio and Brain Engineering, Korea Advanced Institute of Science and Technology (KAIST), Daejeon 34141, Republic of Korea (E-mail: depecher@kaist.ac.kr).}
\thanks{Jong Chul Ye is with the Kim Jaechul Graduate School of AI, KAIST, Daejeon 34141, Republic of Korea (E-mail: jong.ye@kaist.ac.kr).}
}

\maketitle

\begin{abstract}
The widespread application of artificial intelligence in health research is currently hampered by limitations in data availability.
Distributed learning methods such as federated learning (FL) and shared learning (SL) are introduced to solve this problem as well as data management and ownership issues with their different strengths and weaknesses. The recent proposal of federated split task-agnostic (\textsc{FeSTA}) learning tries to reconcile the distinct merits of FL and SL by enabling the multi-task collaboration between participants through Vision Transformer (ViT) architecture, but they suffer from higher communication overhead. To address this, here we present a multi-task distributed learning using ViT with random patch permutation.
Instead of using a CNN based head as in FESTA, p-FESTA adopts a  randomly permuting simple patch embedder, improving the multi-task learning performance without sacrificing
privacy. 
 Experimental results confirm that the proposed method significantly enhances the benefit of multi-task collaboration, communication efficiency, and privacy preservation, shedding light on practical multi-task distributed learning in the field of medical imaging.
\end{abstract}


\section{Introduction}
\label{sec:introduction}
\IEEEPARstart{A}{rtificial} intelligence (AI) has been gaining unprecedented popularity thanks to its potential to revolutionize various fields of data science. Specifically, the deep neural network has attained expert-level performances in the various applications of medical imaging \cite{hosny2018artificial,niazi2019digital}.

To enable the AI models to offer precise decision support with robustness, an enormous amount of data are indispensable. However, data collected from volunteer participation of only a few institutions cannot fully meet the amount to guarantee robust performances. Even for the large public datasets, it may inevitably include unquantifiable biases stemming from the limited geographic regions and patient demographics such as ethnicities and races, resulting in performance instability in  real-world applications. Especially for the newly emerging disease like Coronavirus disease 19 (COVID-19), this limitation can be exacerbated as it is hard to build a large, well-curated dataset with sufficient diversity promptly.

Therefore, the ability to collaborate between multiple institutions is critical for the successful application of AI in medical imaging, but the rigorous regulations and the ethical restrictions for sharing patient data is an another obstacle to multi-institutional collaborative work. Several formal regulations and guidelines, such as the United States Health Insurance Portability and Accountability Act (HIPAA) \cite{edemekong2018health} and the European General Data Protection Regulation (GDPR) \cite{hoofnagle2019european}, state the strict regulations regarding the storage and sharing of patient data.

Accordingly, distributed learning methods, which perform learning tasks at edge devices in a distributed fashion,  can be effectively utilized in healthcare research.  Specifically, distributed learning was introduced to enable the model training with data that reside on the source devices without sharing. Federated learning (FL) is one of these methods that enables distributed clients to collaboratively learn a shared model without sharing their training data \cite{konevcny2016federated}. However, it still holds several limitations in that it is heavily dependent on the client-side computation resources for parallel computation and not completely free from privacy concerns with gradient inversion attack \cite{li2020federated, mammen2021federated}. Another distributed learning method, split learning (SL) \cite{vepakomma2018split}, which splits the network into parts between clients and the server, is a promising method that puts low computational loads at the edge devices; however, it has the disadvantage of high communication overhead between the clients and server \cite{thapa2020splitfed}, and also has limitation in privacy preservation as the private data can be recovered by the malicious attack with feature hijacking and model inversion \cite{gawron2022feature}. In addition, SL show significantly slower convergence compared with FL and shows suboptimal performance under significantly skewed data distribution between clients \cite{gao2020end}.

Inspired by the modular decomposition structure of Vision Transformer (ViT), a novel distributed learning method dubbed \emph{Federated Split Task-Agnostic learning \textsc{(FeSTA)}} was  recently proposed for distributed multi-task collaboration using ViT architecture \cite{park2021federated}. The \textsc{FeSTA} framework, equipped with the shared task-agnostic ViT body on the server-side and multiple task-specific convolutional neural network (CNN) heads and tails on the clients-side, was able to balance the merit of FL and SL, thereby improving the performances of individual tasks under distributed multi-task collaboration setting at a level even better than the single-task expert model trained in a data-centralized manner.

Nevertheless, there remain several critical limitations with the \textsc{FeSTA} framework. First, the communication overhead is higher than that of SL and FL, as the model should continuously share features and gradients as well as head and tail parts of the network, which may impose difficulties in practical implementation. Second, we found that  the large size head and tail parts in the original FeSTA tends to reduce the role of the shared body, resulting in a small improvement compared to the single task learning despite the ViT's the potential for multi-task learning (MTL). Finally, the \textsc{FeSTA} framework was not free from the privacy issue, as the features transmitted to the server body can be hijacked and reverted to the original data by the outside malicious attackers or "honest-but-curious" server in the same manner in SL.

To alleviate these drawbacks, here we introduce \textit{p}-\textsc{FeSTA} framework, a \emph{Federated Split Task-Agnostic learning with permutating pure ViT}, which empowers communication efficient MTL with privacy-preservation. Although the overall composition of \textsc{\textit{p}-FeSTA} is similar to that of \textsc{FesTA}, instead of using a CNN based head, \textit{p}-\textsc{FeSTA} adopts a simple and task non-specific patch embedder like a vanilla ViT, enforcing the self-attention within the transformer architecture to improve the MTL performance.
For privacy preservation, we introduce a \emph{Permutation module} that  randomly shuffles the order of all patch features ahead of sending them to the server, to prevent either outside attacker or "honest but curious" server from reverting features into original data containing privacy.

The new architectural change gives the  \textit{p}-\textsc{FeSTA}  several unique advantages. First, the communication overhead was reduced significantly by saving features to be used throughout the entire learning process.
Furthermore, the benefit of MTL is enhanced by enforcing the head to play a small role and the multi-task body to do heavy lifting.
In addition, data privacy is also enhanced with a simple but effective {permutation module} using the intrinsic property of ViT.

\section{Related Works}

\begin{figure*}[!h]
\centering
\includegraphics[width=0.92\textwidth]{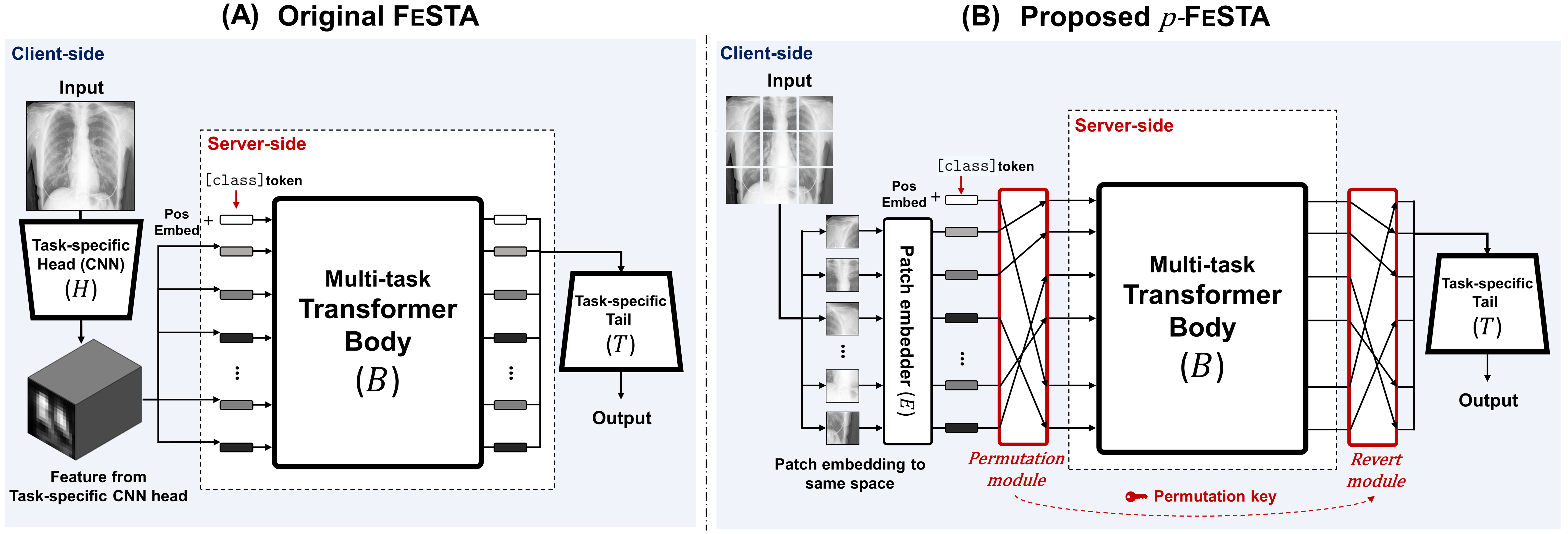}
\caption{
The model configuration of (A) the original \textsc{FeSTA} and (B) the proposed \textsc{\textit{p}-FeSTA} frameworks.
} 
\label{fig1}
\end{figure*}

\begin{figure*}[!h]
\centering
\includegraphics[width=1.0\textwidth]{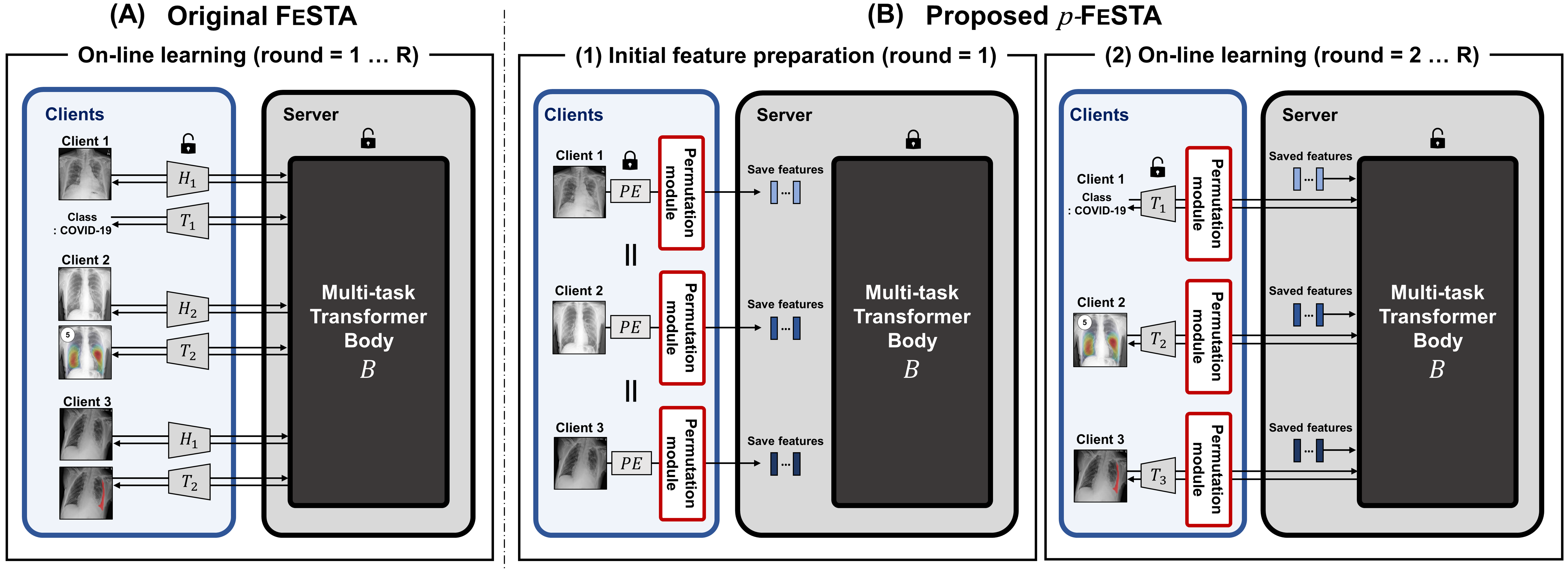}
\caption{
The learning process of (A) the original \textsc{FeSTA} and (B) the proposed \textsc{\textit{p}-FeSTA} frameworks.
} 
\label{fig2}
\end{figure*}

\subsection{Vision Transformer (ViT)}

ViT \cite{dosovitskiy2020image}, a recently introduced deep learning model equipped with an exquisite attention mechanism inspired by its successful application in natural language processing (NLP), has demonstrated impressive performances across many vision tasks. The  multi-head self-attention in ViT can flexibly attend to a sequence of patches of the image to encode the cue, enabling the model to be robust to nuisances like occlusion, spatial permutation, and adversarial perturbation, and thereby having the model to be more shape-biased like human than CNN-based model \cite{naseer2021intriguing}.

In addition, the modular design of the ViT is straightforward, implying that the components can be easily decomposed into parts: head to project the images patches into embeddings, transformer body to encode the embeddings, and tail to yield task-specific output. This easily decomposable design offers the possibility in the application for MTL.
Recall that the motivation of MTL is originated from attempts to mitigate the data insufficiency problem where the numbers of data for individual tasks are limited. MTL can offer the advantage of improving data efficiency, reducing overfitting through shared representation, and faster convergence by leveraging auxiliary knowledge. 

Specifically, MTL with transformer-based models has emerged as a popular approach to improve the performances of the closely related task in NLP \cite{chen2021multi,liu2019multi}. In this approach, a shared transformer learns several related tasks simultaneously, like sentence classification and word prediction, and the tasks-specific module yields the outcome for each task. As shown in previous literature \cite{liu2019multi}, the model trained with MTL strategy generally shows improved performances in a wide range of tasks. Even though not well been studied as in language, the decomposable design of ViT has unleashed the application of MTL to visual transformer models. In an early approach \cite{chen2020pre}, the ViT was divided into the task-specific head, tail, and shared transformer structures across the tasks, and it was possible to attain a similar generalization performance with fewer training steps, by sharing the transformer model among the related tasks.

\subsection{Federated Split Task-Agnostic (\textsc{FeSTA}) Learning}
The main motivation of existing \textsc{FeSTA} framework as described in Fig.~\ref{fig1}A and Fig.~\ref{fig2}B was to devise a framework to maximally exploit the distinct strengths of FL and SL methods and to improve the performances of individual tasks with collaboration between clients performing various tasks. 
 
Let $\Cc=\bigcup_{k=1}^K C_k $ be a group of client sets with different tasks, where $K$ denotes the number of tasks and the client set $C_k$ includes one or more clients having different data sources for the $k$-th task, i.e. $C_k=\{c_{1}^{k}, c_{2}^{k}, \dots, c_{N_k}^{k} : N_k \geq 1 \}$. Clients in each client set for the $k$-th task has its own task-specific model architecture for head $\Hc_c$ and a tail $\Tc_c$, while the server-side transformer body $\Bc$ is shared.

For training, the server and each client initialize the weights of each sub-network with random initialization or from the pre-trained parameters. For learning round {$i = 1,2,\ldots R$}, individual clients do the forward pass on their task-specific head $\Hc_c$ using the local training data $\{(x_c^{(i)},y_c^{(i)})\}_{i=1}^{N_c}$, and send the intermediate feature $h_c^{(i)}$ to the server: $h_c^{(i)} = \Hc_c(x_c^{(i)})$. 
The transformer body $\Bc$, then receives the intermediate features from all clients and gets features $b_c^{(i)}$ in parallel with the forward pass, to send them back to each client $c$: $b_c^{(i)} = \Bc(h_c^{(i)})$.
With the features $b_c^{(i)}$, the task-specific tail in client yield the output $\hat y_c^{(i)}=\Tc_c(b_c^{(i)})$, and forward pass finishes.
Back-propagation is performed exactly the opposite way, in order of tail, body, and head. First, loss is calculated in tail as: $\ell_c(y_c^{(i)}, \Tc_c(\Bc(\Hc_c(x_c^{(i)}))))$, where $\ell_c(y,\bar y)$ denotes the $c$ task-specific loss between the target $y$ and the estimate $\bar y$. Then, the gradients are pass from tail, body to head in reverse order to forward-propagation, using the chain-rule.

For multi-task body update, the optimization is performed by fixing the head and tails.
For the task-specific head and tail updates, the optimization problem is solved by fixing the Transformer body.
In addition, per every "UnifyingRounds", the server aggregates, averages and distributes the head and tail parameters between clients participating in the same task, as in FedAvg \cite{mcmahan2017communication}.

In the previous study, the \textsc{FeSTA} along with the MTL was shown to ameliorate the individual performances of the clients in collaboration, while resolving the data governance and ownership issue as well as eliminating the need to transmit the huge weights of the transformer body \cite{park2021federated}.

\section{Method}

\subsection{\textsc{\textit{p}-FeSTA}}

Nonetheless, the \textsc{FeSTA} framework still has several drawbacks. First, the communication cost can be  higher since the features and gradients should be continuously exchanged between the server and clients like in SL but the head and tail weights should also be aggregated between the clients as in FL.
Accordingly, the total communication costs are inevitably higher than SL, and even higher than FL depending on the network size. Second, as shown in our ablation study without the transformer body, the CNN head and tail themselves already have strong representation capacity, which may diminish the role of the transformer body between head and tail. Third, privacy concerns may arise as there is no privacy-preserving method from the model inversion attack on the feature transmitted from client to server.

\begin{figure*}[!h]
\centering
\includegraphics[width=0.8\textwidth]{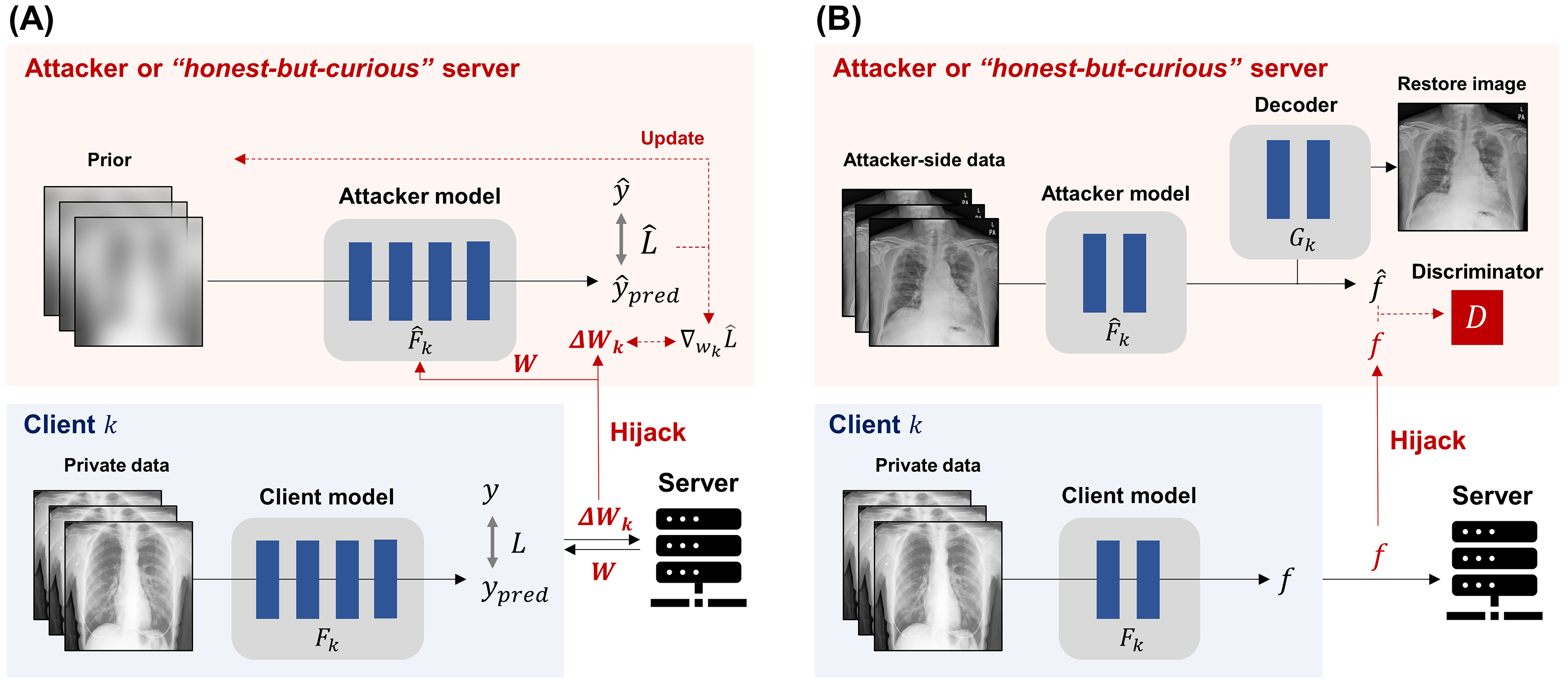}
\caption{Possible privacy attacks in (A) federated learning and (B) split learning.} 
\label{fig3}
\end{figure*}

\begin{figure*}[!h]
\centering
\includegraphics[width=0.93\textwidth]{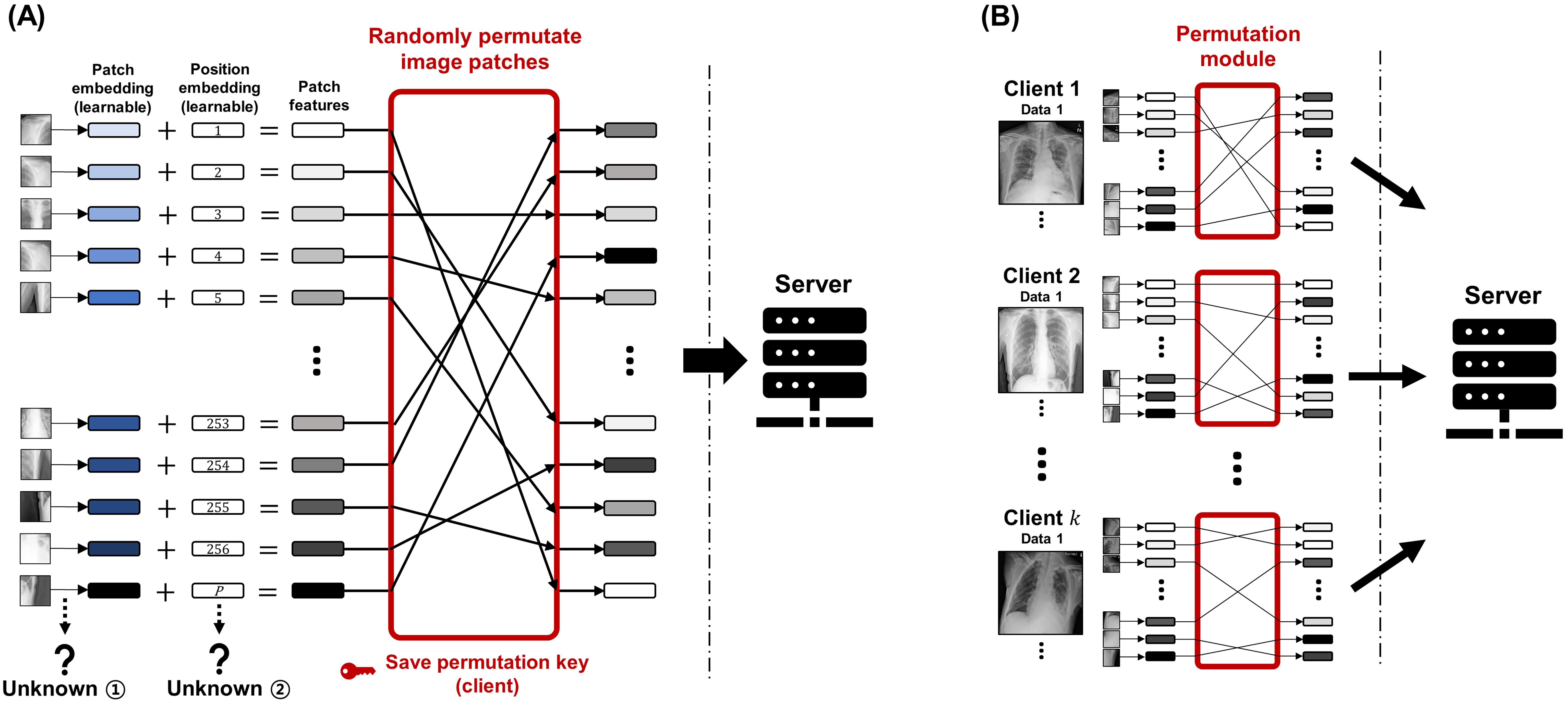}
\caption{
(A) The \emph{Permutation module} to enhance privacy and (B) the illustration of different permutation patterns for each data of each client. 
} 
\label{fig4}
\end{figure*}

The proposed \textsc{\textit{p}-FeSTA} is a framework devised to mitigate these shortcomings.
As shown in Fig.~\ref{fig1}B and Fig.~\ref{fig2}B,
the overall composition of \textsc{\textit{p}-FeSTA} is similar to that of \textsc{FesTA}, which decomposes networks into head $\Hc$, body $\Bc$ and tail $\Tc$. However, unlike previous \textsc{FeSTA}, we do not use the CNN head tailored for each task. Having the CNN head to be powerful enough to play a major role in the task hinders the shared transformer from being an important component as there remains little room to improve with this additional module. Instead, we adopted a simple and task non-specific patch embedder like a vanilla ViT, enforcing the self-attention within the transformer architecture to do the heavy lift. 

Unfortunately, the use of patch embedding in a vanilla ViT may  be prone to outside attackers that attempts to invert the patch embedder to obtain the original images. To address this,
here we propose a novel \emph{permutation module} as depicted in Fig.~\ref{fig1}B to prevent either outside attacker or "honest but curious" server from reverting features into original data containing privacy. Specifically, this \emph{Permutation module} randomly shuffles the order of all patch features ahead of sending them to the server, and stores the key to reverse the permutation on the client-side. Then, the transformer body $\Bc$ in the server does a forward pass with the permutated patch features and sends the encoded features back to the clients. Finally, the client reverses the permutation with the saved key and yields the final output by passing the reverted features to the task-specific tail $\Tc_k$. The back-propagation is performed in the exact opposite way, where the same \emph{Permutation module} to forward-propagation is utilized.

The availability of \emph{Permutation module} attributes to an intriguing property of ViT that all the components composing the transformer body, such as multi-head self-attention, feed-forward network, and layer normalization, is fundamentally "permutation equivariant" \cite{naseer2021intriguing}. They are processed independently in a patch-based manner and the order of the patch does not affect the outcome, and therefore, the transformer body can be trained without any performance degradation. In addition, as the orders of patches are completely shuffled, it is infeasible for a malicious attacker to successfully revert the original image. 
How the \emph{Permutation module} can provide privacy protection from the malicious attacker will be described in more detail in the following.

\subsection{Protecting Privacy with Permutation Module}

For FL, privacy is improved by the ephemeral and focused nature of the federated aggregation, averaging, and distribution of the model updates, assuming that the model updates are considered to be less informative than the original data. However, recent studies have thrown doubt to the false sense of security, showing that private data can be uncovered faithfully only with these local model updates \cite{geiping2020inverting, zhu2019deep, zhao2020idlg}. In detail, given the access to the global model $W$ and the client's model update $\Delta W$, the attacker can optimize the input image from a prior to produce a gradient that matches the client's model update as illustrated in Fig.~\ref{fig3}A. However, this type of attack is infeasible for the proposed \textsc{\textit{p}-FeSTA} method, since only the tail part of the entire model is aggregated and distributed by the server to the clients. For instance, for COVID-19 classification, the task-specific tail is a simple linear classifier, with which the original data with privacy cannot be uncovered.

SL protects privacy in a different way. As the name suggests, SL split the entire model into client-side and server-side sub-networks and does not send the models between the server and clients.
Instead, the features and gradients are transmitted back and forth between the server and clients, and it can be the prey of the malicious attacker \cite{gawron2022feature}. As described in Fig.~\ref{fig3}B, when clients send the intermediate features $f$ to the server, the attackers may hijack the features, and instead of running the remainder of the SL model, they train three components: encoder $\hat{F}$, decoder $G$, and discriminator $D$ with their own data. The $D$ is trained to discriminate between the hijacked feature $f$ and the feature $\hat{f}$ encoded by $\hat{F}$, which enforces $\hat{f}$ to be in the same feature space to $f$. Simultaneously, $G$ learns to decode $\hat{f}$ into the image with minimal error. Then, well-trained $G$ can also be used to decode the hijacked feature $f$ to original data faithfully.

The feature space hijacking is also possible for our \textsc{\textit{p}-FeSTA}. To make the matter worse, the head part of our model is relatively simple and can be easy prey for an attacker. This is why we introduce a novel \emph{Permutation module} to protect privacy as described in Fig.~\ref{fig4}A. The \emph{Permutation module} randomly shuffles the order of all patch features. 
{In the implementation, the permutations of each data of each client are all different without any regularity as shown in Fig.~\ref{fig4}B, resulting in innumerable patterns for all data.}
With these random permutations, even if a malicious attacker or server steals patch features to uncover private data, the parameter of position embedding, an unknown learnable variable, cannot be inferred as they have no information about the original order of patches. It is also infeasible to inverse the patch features to image patches since the added position embedding, which is unknown to the attacker, should be subtracted first for inversion. This makes a contradiction that the attacker should already know an "unknown" to infer the other "unknown", making the inversion attack a type of underdetermined problem.

\subsection{Training Procedure}
The learning process of the \textsc{\textit{p}-FeSTA} is akin to the original \textsc{FeSTA}, but dissimilar in several aspects. Instead of task-specific head $\Hc_k$ for each task $k$, the task non-specific patch embedder $\Hc$ prepares the patch embeddings $h_c$ for each client $c$ at the beginning and sends them to the server after passing them through the \emph{Permutation module}. The server then saves the received patch embeddings $h_c$ on its side and uses them throughout the remainder of the learning process in order to update the body $\Bc$ and tail $\Tc_k$ parts of the model. Consequently, the overall communication costs can be
significantly reduced compared to the original \textsc{FeSTA}, as the communications to send the intermediate feature $h_c$ or to update the head $\Hc$ are no more required.

As can be seen, the head part of the model, the patch embedder, cannot be updated in this configuration. However, fixing the parameters of the patch embedder did not bring about any performance exacerbation thanks to the simple structure to just embed the image patches into the same vector spaces. Having them trainable rather slightly decreased the performances by resulting in the discrepancy in embedding between tasks. The experimental results will be provided in the ablation study of Section III-H. The detailed process of the proposed \textsc{\textit{p}-FeSTA} is formally described in Algorithm~\ref{alg:algo2}.

\begin{algorithm}[h]
    \caption{Proposed \textsc{\textit{p}-FeSTA} algorithm}
    \label{alg:algo2}
    \DontPrintSemicolon
    \SetKwProg{Fn}{Function}{:}{}
    \SetKwFunction{ServerMain}{ServerMain}
    \SetKwFunction{ClientHead}{ClientHead}
    \SetKwFunction{ClientTail}{ClientTail}
    \SetKwFunction{ClientUpdate}{ClientUpdate}
    \SetKwFor{ForP}{for}{do in parallel}{endfor}
    \SetAlgoNoLine
    \Fn{\ServerMain}{
        \SetAlgoVlined
        Initialize server body weight client head/tail weights
        \ForP{$\mathbf{tasks} \ k \in \{1,2,\ldots K\}$}{
                \ForP{$\mathbf{clients} \ c \in C_k$}{
                $ h_{c} \leftarrow \ClientHead(c)$ \;
                Save patch embedding $h_{c}$ in server memory
                }
        }
        
        \For{$\mathbf{rounds} \ i = 1,2,\ldots R$}{
            \ForP{$\mathbf{tasks} \ k \in \{1,2,\ldots K\}$}{
                \ForP{$\mathbf{clients} \ c \in C_k$}{
                    \If{$i = 1 \ \mathbf{or} \ (i-1) \in \textnormal{UnifyingRounds}$}{
                     Set client $ {{w}}_{\Tc_c}^{(i)} \leftarrow {{\bar{w}}_{\Tc, k}}$  \;
                    }
                    Load $ h_{c}^{(i)}$ by batch from server memory \& $b_{c}^{(i)} \leftarrow \Bc( h_{c}^{(i)})$ \;   
        			${{\partial{L}_{c}^{(i)}}\over  {\partial b_{c}^{(i)}}}  \leftarrow \ClientTail(c,b_{c}^{(i)})$ \& Backprop.\;
                    ${{w}}_{\Tc_c}^{(i+1)} \leftarrow \ClientUpdate(c)$ \;
                }
            }
            Update body ${{w}}_\Bc^{(i+1)} \leftarrow {{w}}_{\Bc}^{(i)} - {\eta \over {K}}\sum\limits^{K}_{k=1}\sum\limits_{c \in C_k}{{\partial{L}_{c}^{(i)}}\over N_k{\partial{w}_\Bc^{(i)}} }$ \;
            \If{$i \in \textnormal{UnifyingRounds}$}{
                \For{$\mathbf{tasks} \ k \in \{1,2,\ldots K\}$}{
                     Update ${\bar{w}}_{\Tc, k} \leftarrow {1 \over N_k}\sum\limits_{c \in C_k}{{w}}_{\Tc_c}^{(i+1)}$ \;
                    }
                }
        }
    }
    \Fn{\ClientHead{$c$}}{
        $ x_c \leftarrow$ All data on client $c$ \;
        $h_c \leftarrow \Hc( x_c)$ \;
        Randomly permutate patch embedding $h_c$ \;
        \KwRet $h_c$\;
    }
    \Fn{\ClientTail{$c,b_c$}}{
        $ y_c \leftarrow$ Current batch of label from client $c$\;
        $ L_c \leftarrow {\ell}_c(y_c, \Tc_{c}(b_c)) $ \& Backprop. \;
        \KwRet ${\partial L_c \over  {\partial b_{c}}}$\;
    }
    \Fn{\ClientUpdate{$c$}}{
    	Backprop. tail, body \& $ {{w}}_{\Tc_c} \leftarrow {{w}}_{\Tc_c} - \eta {{\partial{L_{c}}}\over {\partial{w}}_{\Tc_c}}$ \;
        \KwRet  ${{w}}_{\Tc_c}$\;
    }
\end{algorithm}

\begin{table}[ht]
  \centering
  \caption{Data partitioning for COVID-19 classification}
    \begin{tabular}{cccc}
    \toprule
    \multirow{2}[2]{*}{\textbf{Class}} & \textbf{CNUH}  & \textbf{KNUH}  & \textbf{BIMCV} \\
          & \textbf{(Test)} & \textbf{(Client \#1)} & \textbf{(Client \#2)} \\
    \midrule
    Normal & 417   & 400   & 93 \\
    Other infection & 58    & 400   & - \\
    COVID-19 & 81    & 293   & 782 \\
    \midrule
    Total & 556   & 1093  & 875 \\
    \bottomrule
    \end{tabular}
  \label{tab:1}
\end{table}

\begin{table}[ht]
  \centering
  \caption{Data partitioning for COVID-19 severity prediction}
    \begin{tabular}{cccc}
    \toprule
    \multirow{2}[2]{*}{\textbf{Severity}} & \textbf{CNUH}  & \textbf{YNU}  & \textbf{Brixia} \\
          & \textbf{(Test)} & \textbf{(Client \#3)} & \textbf{(Client \#4)} \\
    \midrule
    1     & 26    & 63    & 261 \\
    2     & 11    & 59    & 443 \\
    3     & 8     & 25    & 414 \\
    4     & 7     & 35    & 866 \\
    5     & 12    & 18    & 745 \\
    6     & 17    & 86    & 1536 \\
    \midrule
    Total & 81    & 286   & 4265 \\
    \bottomrule
    \end{tabular}
  \label{tab:2}
\end{table}

\begin{table}[ht]
  \centering
  \caption{Data partitioning for pneumothorax segmentation}
    \begin{tabular}{cccc}
    \toprule
    \multirow{2}[2]{*}{\textbf{Data}} & \textbf{Subset \#1}  & \textbf{Subset \#2}  & \textbf{Subset \#3} \\
          & \textbf{(Test)} & \textbf{(Client \#5)} & \textbf{(Client \#6)} \\
    \midrule
    Total & 1000  & 4840  & 4839 \\
    \bottomrule
    \end{tabular}
  \label{tab:3}
\end{table}

\section{Results}
\subsection{Implementation Details}
As for the head part, we used the task non-specific patch embedder consisting of the convolution layer with a kernel size of $16\times16$ and stride of 16, input channel of 3, and output channel of 768. For the server-side body, the transformer encoder of the ViT-base model, consisting of 12 encoder layers and 12 attention heads, was used. For the tail part, the network architectures specialized to yield the task-specific output were adopted. For the COVID-19 classification task, we used a simpler linear classifier. For severity prediction, the mapping module with five up-sizing convolution layers was adopted as proposed in \cite{park2022multi}. For pneumothorax segmentation, the decoder part of U-Net \cite{ronneberger2015u} was used.

We simulated the distributed MTL between the institutions participating in three different CXR tasks: classification, severity prediction of COVID-19, and pneumothorax segmentation.
As in  \cite{park2021federated}, the model was first initialized with pre-trained weights for the CheXpert dataset. We minimized the binary cross-entropy (BCE) losses  for each class for the classification task. The severity of COVID-19 was predicted and evaluated in an array-based manner as suggested in \cite{toussie2020clinical}. Specifically, BCE losses for each six location arrays of the lung were used for the optimization in severity prediction. Finally, for the pneumothorax segmentation, we minimized the binary cross-entropy loss combined with dice and focal losses. The SGD optimizer was used for the classification and severity prediction tasks, while the Adam optimizer was utilized for the segmentation task, with a learning rate of 0.0001 and a warm-up constant learning rate scheduler for all tasks. The batch size was 4 per client, and the warm-up step was 500. The total training round was 6,000 for all clients, and the tail weights are averaged every 100 local iterations.
To adjust the scale of gradients, the 1:2:10 gradient scaling was applied for classification, severity prediction, and segmentation, respectively.

The FL, SL, \textsc{FeSTA} and \textsc{\textit{p}-FeSTA} was simulated on the modification of Flower (licensed under an Apache-2.0 license) \cite{beutel2020flower} framework. All experiments were performed with Python 3.8 and Pytorch 1.8 on Nvidia RTX 3090, 2080 Ti.

\subsection{Practical Simulation for Multi-task Collaboration}
One of the paramount motivations for FL in medical imaging is to make a robust model leveraging the dispersed and small-sized datasets from multiple institutions while avoiding data governance. Therefore, we assume the FL scenario in which the data of several clients are scanty.

For COVID-19 classification and severity prediction, we used both publicly available datasets and private data collected from local institutions. Overall, 1093 CXR from a local hospital (KNUH, client \#1) and 875 from public data (BIMCV \cite{de2020bimcv}, client \#2) were used for training and 556 CXRs from another hospital (CNUH) were used as the external test set in COVID-19 classification task as shown in Table~\ref{tab:1}. Similarly, for the COVID-19 severity prediction task, 286 CXRs from a local hospital (YNU, client \#3) and 4,265 from public data (Brixia \cite{signoroni2021bs}, client \#4) were used as the training, and 81 CXRs data of another hospital (CNUH) were used as the external test set as provided in Table~\ref{tab:2}. For pneumothorax segmentation, we used the Society for Imaging Informatics in Medicine and the American College of Radiology (SIIM-ACR) Pneumothorax Segmentation Challenge \cite{siim2018pneumothorax} dataset consisting of 10,679 CXR images. The randomly selected 1,000 CXR images were used as the test set, and the remaining 9679 CXR images were randomly split with a 1:1 ratio (4840 and 4839 CXRs) into two clients (client \#5 and client \#6) to emulate the participation of two hospitals as in Table~\ref{tab:3}. For practical simulation of collaboration between hospitals, we allocated non-overlapping data sources to each client except for the pneumothorax segmentation task where the exact sources of the data can not be estimated. Overall, six clients participated in the MTL scenario, two clients per task. For this study, Institutional Review Board approvals of each participating hospital were obtained and informed consent was waived. 

Considering the sizes and compositions of each client, collaboration for the COVID-19 classification task can be regarded as the collaboration between all clients having small data with a substantial imbalance in data distribution. Likewise, the collaboration for COVID-19 severity can be considered to be the simulation of an imbalance in data size between the participants, one client has scanty data while the other client has relatively sufficient data, in addition to the differences in data composition. Finally, the clients for pneumothorax segmentation emulate the situation in which each participating clients have relatively sufficient and homogeneous data with similar sizes.

When viewed in terms of the relevance between tasks, the COVID-19 classification and severity prediction task can be considered to be highly correlated tasks, while the pneumothorax segmentation task may be regarded as a less relevant task. 

\begin{table*}[ht]
  \centering
  \caption{Performance comparison with other distributed learning methods}
    \begin{tabular}{ccccccc}
    \toprule
    \multirow{3}[4]{*}{\textbf{Methods}} & \multicolumn{4}{c}{\textbf{Classification}} & \textbf{Severity} & \textbf{Segmentation} \\
          & \multicolumn{4}{c}{\textbf{AUC}} & \multirow{2}[3]{*}{\textbf{MSE}} & \multirow{2}[3]{*}{\textbf{Dice}} \\
\cmidrule{2-5}          & \textbf{Average} & \textbf{Normal} & \textbf{Others} & \textbf{COVID-19} &       &  \\
    \midrule
    Data centralized & 0.671 (0.051) & 0.735 (0.071) & 0.777 (0.045) & 0.500 (0.051) & 1.592 (0.081) & 0.793 (0.005) \\
    Federated learning & 0.601 (0.036) & 0.597 (0.146) & 0.483 (0.068) & 0.722 (0.023) & 2.159 (0.188) & 0.789 (0.001) \\
    Split learning & 0.546 (0.024) & 0.522 (0.067) & 0.534 (0.050) & 0.583 (0.013) & 2.546 (0.414) & 0.790 (0.000) \\
    FeSTA (STL) & 0.718 (0.047) & 0.680 (0.088) & 0.677 (0.032) & 0.795 (0.036) & \textbf{1.318 (0.125)} & 0.801 (0.011) \\
    p-FeSTA (STL) & 0.696 (0.022) & 0.739 (0.093) & 0.557 (0.118) & 0.790 (0.045) & 1.717 (0.148) & 0.803 (0.004) \\
    FeSTA (MTL) & 0.780 (0.019) & 0.785 (0.009) & 0.793 (0.100) & 0.761 (0.034) & 1.416 (0.048) & 0.796 (0.013) \\
    \textbf{p-FeSTA (MTL)} & \textbf{0.884 (0.008)} & \textbf{0.906 (0.004)} & \textbf{0.890 (0.011)} & \textbf{0.857 (0.014)} & 1.361 (0.057) & \textbf{0.808 (0.003)} \\
    \bottomrule
    \multicolumn{5}{l}{\footnotesize  Values are presented in mean (standard deviation) of three repeats with different seed.}
    \end{tabular}%
  \label{tab:4}%
\end{table*}%

\subsection{Performance Metrics}
To evaluate the classification performance, the area under the receiver operating characteristic curve (AUC) was used. For the severity prediction task, the mean squared error (MSE) of prediction was used as in the previous work \cite{park2022multi}. To evaluate the segmentation accuracy, the Dice coefficient was calculated to measure the intersection of the segmentation results and ground truth annotations. All experiments were performed repeatedly with three different seeds to exclude the coincidence of getting over- or underestimated results.

\subsection{Comparison Results}
Table~\ref{tab:4} shows a comparison of the proposed \textsc{\textit{p}-FeSTA} with data centralized training, other distributed learning, and original \textsc{FeSTA} methods. For a fair comparison, all other methods underwent the same pre-training step as the proposed method. The same model architectures were used for all other distributed learning methods except for the original \textsc{FeSTA} methods, in which the task-specific CNN head is a key part of the method. For original \textsc{FeSTA} methods, DenseNet-121 equipped with PCAM operation \cite{ye2020weakly} tailored for CXR classification were used as the head instead of the simple patch embedder as in our previous work \cite{park2022multi}.

The single-task models trained with either \textsc{\textit{p}-FeSTA} showed a similar order of magnitude in performances of three tasks compared with data centralized training, surpassing those of FL and SL. The improvements with the \textsc{\textit{p}-FeSTA} over FL and SL were noticeable, especially for the classification and severity prediction tasks, where the data insufficiency and imbalance problems are prominent. Note that the slightly better performance of the single-task model trained with the original \textsc{FeSTA} than the \textsc{\textit{p}-FeSTA} for the task of severity quantifiacation, which is even better than data centralized learning, may attribute to the more expressive task-specific CNN head tailored for CXR tasks.

As shown in Table~\ref{tab:4}, the model obtained with MTL between three tasks using \textsc{\textit{p}-FeSTA} significantly outperforms the single-task counterparts and all other distributed learning methods.
Note that the performance gain with the MTL over the single-task model is more prominent in the \textsc{\textit{p}-FeSTA} than the previous \textsc{FeSTA}. Even when compared with the MTL model obtained with \textsc{FeSTA}, \textsc{\textit{p}-FeSTA} showed similar or slightly better performance in severity prediction and segmentation, but substantially outperformed the previous one in the classification task, providing generally superior performances. 

The fact that the benefit of MTL is formidable in classification and severity prediction is intriguing, as they are the tasks in which scanty data with skewed distribution are problematic. On the contrary, the performance improvement was modest for pneumothorax segmentation where each participating clients have a relatively large number of data with even distribution. Moreover, the close relevance between COVID-19 classification and severity prediction might have further enhanced the benefit of MTL to those tasks, compared with the relatively less relevant task of pneumothorax segmentation.

\subsection{Communication Costs between Server and Clients}
In this section, we provide the estimated communication costs between the server and clients. Given the number of data as $D$, the batch size as $B$, the rounds between aggregation and distribution by the server as $n$, and the transmission of features, gradients, and the head, body, tail parameters as $F$, $G$, and $P_{h}$, $P_{b}$, $P_{t}$, the communication costs $T$ of each distributed learning strategies for a total of $R$ rounds between the server and one client can be formulated as follow:
\begin{gather}
   T_\text{FL} =  {2R \over n}(P_h + P_b + P_t), \\
   T_\text{SL} = 2BR (F + G) , \\
   T_\text{\textsc{FeSTA}} = 2BR (F + G) + {2R \over n}(P_h + P_t), \\
   T_\text{\textsc{\textit{p}-FeSTA}} = {DF} + BR ({F} + {G}) + {2R{P_t} \over n},
\end{gather}
where the constant 2 is multiplied to take account of the both-way transmissions between server and client. Note that the cost for features and gradients transmissions are not multiplied by 2 in \textsc{\textit{p}-FeSTA} to reflect no transmission of features and gradients to the head during the learning process.

\begin{table}[h]
  \centering
  \caption{Communication costs of the distributed learning methods}
    \begin{tabular}{cccc}
    \toprule
          & \textbf{Total} & \textbf{Features/gradients} & \textbf{Parameters} \\
    \midrule
    \textbf{Classification} &       &       &  \\
    FL    & 10456.152M & -     & 10456.152M  \\
    SL    & 9474.048M  & 9474.048M  & - \\
    \textsc{FeSTA} & 11390.423M  & 9474.048M  & 1916.375M  \\
    \textsc{\textit{p}-FeSTA} & 4880.648M  & 4844.890M  & 35.758M  \\
    \midrule
    \textbf{Severity} &       &       &  \\
    FL    & 11090.794M  & -     & 11090.794M  \\
    SL    & 9474.048M  & 9474.048M  & - \\
    \textsc{FeSTA} & 12025.065M  & 9474.048M  & 2551.017M  \\
    \textsc{\textit{p}-FeSTA} & 5435.649M  & 4765.249M  & 670.401M  \\
    \midrule
    \textbf{Segmentation} &       &       &  \\
    FL    & 11160.985M  & -     & 11160.985M  \\
    SL    & 9474.048M  & 9474.048M  & - \\
    \textsc{FeSTA} & 12095.256M  & 9474.048M  & 2621.208M  \\
    \textsc{\textit{p}-FeSTA} & 5899.113M  & 5158.520M  & 740.592M  \\
    \bottomrule
    \end{tabular}%
  \label{tab:5}%
\end{table}%

\begin{table*}[ht]
  \centering
  \caption{Ablation studies for the proposed \textit{p}-FeSTA}
    \begin{tabular}{ccccccc}
    \toprule
    \multirow{3}[4]{*}{\textbf{Methods}} & \multicolumn{4}{c}{\textbf{Classification}} & \textbf{Severity} & \textbf{Segmentation} \\
          & \multicolumn{4}{c}{\textbf{AUC}} & \multirow{2}[3]{*}{\textbf{MSE}} & \multirow{2}[3]{*}{\textbf{Dice}} \\
\cmidrule{2-5}          & \textbf{Average} & \textbf{Normal} & \textbf{Others} & \textbf{COVID-19} &       &  \\
    \midrule
    Proposed & 0.884 (0.008) & 0.906 (0.004) & 0.890 (0.011) & 0.857 (0.014) & \textbf{1.361 (0.057)} & 0.808 (0.003) \\
    w learnable head & 0.890 (0.001) & 0.909 (0.014) & 0.895 (0.005) & \textbf{0.866 (0.013)} & 1.545 (0.386) & 0.789 (0.000) \\
    w/o permutation & \textbf{0.890 (0.010)} & \textbf{0.909 (0.002)} & \textbf{0.904 (0.023)} & 0.858 (0.008) & 1.461 (0.064) & \textbf{0.809 (0.002)} \\
    w/o position embedding & 0.827 (0.028) & 0.831 (0.035) & 0.786 (0.049) & 0.862 (0.007) & 1.942 (0.112) & 0.798 (0.004) \\
    \bottomrule
    \multicolumn{5}{l}{\footnotesize  Values are presented in mean (standard deviation) of three repeats with different seed.}
    \end{tabular}%
  \label{tab:6}%
\end{table*}%

Numerically, the communication cost for each distributed learning method in our experimental setting can be calculated as in Table~\ref{tab:5}.
As one of the critical drawbacks, the communication cost of \textsc{FeSTA} is  larger than SL and even higher than FL.
On the contrary, the proposed \textsc{\textit{p}-FeSTA} substantially lessens the communication burden by saving the head features at the beginning and using them throughout the entire learning process on the server-side. In our experimental setting, the total communication overhead of the proposed \textsc{\textit{p}-FeSTA} is less than half of the previous \textsc{FeSTA}, and also significantly lower than SL as well as FL.

\subsection{Ablation Studies}
Results of ablation studies are suggested in Table~\ref{tab:6}. 

\subsubsection{Fixed Head}
We first performed an ablation to verify whether fixing the head parametersand the patch embedder does not harm the performance. Compared with the proposed method with a fixed head, the same model trained using the learnable head showed similar or even slightly worse performance for severity prediction and segmentation, which may attribute to the overfitting to training data. Therefore, we concluded that the patch embedder can be fixed during all the learning rounds without concerns of performance degradation.

\subsubsection{Permutation Module}
We next ablated the \emph{Permutation module} to verify whether our method is indeed "permutation equivariant". For this proposition to be true, the performance should be the same regardless of the presence of \emph{Permutation module}. As expected, the performances with and without the \emph{Permutation module} were in the same order of magnitude, with the differences falling within standard deviations, proving that the permutation does not affect the performance of the transformer model.

\subsubsection{Position Embedding}
In the proposed method, the position embedding takes two roles, first provides position information to yield the final output in the tail, and second adds an unknown parameter to prevent an attacker from uncovering patch features into image patches. We performed the ablation study to confirm that the position embedding is necessary for optimal performance in addition to privacy preservation. The model trained without the position embedding showed slightly lower performances than that with the position embedding in all tasks, suggesting that the position embedding is indispensable for the best performance as well as privacy preservation.

\section{Discussion}
In this work, we introduced a significantly improved federated task-agnostic learning framework with permutating pure ViT, dubbed \textsc{\textit{p}-FeSTA}, which resolves the major drawbacks of our previous \textsc{FeSTA} framework, leveraging the intrinsic properties of the ViT. The newly proposed \textsc{\textit{p}-FeSTA} substantially reduces the communication overhead between server and clients as well as enhances the performance with the authentic multi-task training in the same embedding space, while offering better privacy preservation. Table~\ref{tab:7} summarizes the comparison between the proposed \textsc{\textit{p}-FeSTA}, original \textsc{FeSTA} and other distributed learning methods.

\begin{table}[t]
  \centering
  \caption{Comparison of the distributed learning methods.}
    \begin{tabular}{ccccc}
          & \textbf{FL} & \textbf{SL} & \textbf{\textsc{FeSTA}} & \textbf{\textsc{\textit{p}-FeSTA}} \\
    \midrule
    Model averaging & O     & X     & O     & O \\
    Client-side learning & Parallel & Sequential & Parallel & Parallel \\
    Model split & X     & O     & O     & O \\
    Communication cost & High & High  & High & Low \\
    Benefit of MTL & X     & X     & Small & Large \\
    Privacy protection & X     & X     & X     & O \\
    \end{tabular}%
  \label{tab:7}%
\end{table}%

One of the most tackling problems of the previous \textsc{FeSTA} method was the  communication overhead between server and clients since it requires the feature and gradient transmission the same as in SL as well as the server-side aggregation and distribution of the heads and tails parameters for each client. This configuration increases the communication cost to be inevitably larger than SL and even larger than FL according to the network sizes of each model component. To mitigate the problem, we configured the head part to be a simpler structure like a patch embedder so that the pre-trained head can sufficiently show the best performance without the additional training of the head that requires communications back and forth between server and clients. Consequently, the feature from the head could be stored on the server-side at the beginning and used throughout the entire learning process, reducing the overall communication cost to approximately half of other distributed learning methods.

Moreover, having the head part be a common patch embedder provides another advantage of embedding the image features of different tasks to be in the same embedding spaces.
This results in the increasing role of following multi-task transformer and facilitates learning better shared representation, compared with our previous method where task-specific CNN heads embed the features into different embedding spaces for each task and confine the role of transformer resultingly.

Concerns may arise that using a simpler head structure and saving features in the server-side memory could unleash privacy leakage, which is especially important for medical data. To alleviate this concern, we utilized the intriguing properties of the ViT, the "permutation equivariance" of the self-attention mechanism. The patch features were randomly permutated ahead of transmission to the server, making the problem underdetermined to the attacker while not deteriorating the performance of the transformer.

The merit of our method can be maximized in "data-hungry" collaboration. As shown in the experiments, the performance gain was more prominent for the classification task and severity prediction tasks where the data of each client are scanty. Given that one of the important motivations of distributed learning is to enable building a robust model without data centralization with the participation of many clients having limited data, this potential gain in data-hungry collaboration will further incentivize the widespread application.

Nevertheless, our study is not free of limitations. First, even though we simulated the practical collaboration between hospitals on the customized Flower framework, the robustness to the other tackling factor such as straggler-resilience was not verified \cite{li2020federated, reisizadeh2020straggler}. Considering that connection instability becomes a common problem in online learning, it should be resolved technically ahead of real-world implementation. Second, we did not consider other types of attacks for distributed learning, such as model poisoning or data poisoning \cite{lyu2020threats, tolpegin2020data, bagdasaryan2020backdoor}, which is beyond the scope of this work. For defense against these types of malicious attacks, the existing methods \cite{lyu2020privacy, zhao2019pdgan, li2020learning} can be utilized along with our framework. Future work might verify the robustness for these types of malicious attacks.

\section{Conclusion}
In this paper, we proposed the novel \textsc{\textit{p}-FeSTA} framework with pure ViT, which elicits the synergy of MTL among heterogeneous tasks as well as reduces the communication overhead significantly compared to the existing FeSTA.
In addition, we also enhanced the privacy using the \emph{Permutation module} in a way specific to ViT. We believe that our work is a step toward facilitating distributed learning among the institutions wanting to participate in different tasks, mitigating the major drawbacks of the existing methods.

\section*{Acknowledgement}
This research was supported by a grant of the MD-Phd/Medical Scientist Training Program through the Korea Health Industry Development Institute (KHIDI), funded by the Ministry of Health \& Welfare, Republic of Korea. {We are grateful to Gwanghyun Kim (Seoul National University) for his help in the implementation of the framework for the experiments.}

\bibliographystyle{IEEEtran}
\bibliography{fed}

\end{document}